\documentclass{egpubl}
\usepackage{egsgp2022}
\JournalPaper         

\usepackage[T1]{fontenc}
\usepackage{dfadobe}  

\usepackage{cite}  
\BibtexOrBiblatex
\electronicVersion
\PrintedOrElectronic
\ifpdf \usepackage[pdftex]{graphicx} \pdfcompresslevel=9
\else \usepackage[dvips]{graphicx} \fi

\usepackage{egweblnk} 

\usepackage{enumitem}
\usepackage{amsmath}
\usepackage{amsfonts}
\usepackage{multirow}
\usepackage{bigdelim}
\usepackage{color}
\usepackage{booktabs}
\usepackage{ifthen}
\usepackage{hyperref}
\usepackage{subcaption}
\usepackage{bbold}
\usepackage[noabbrev]{cleveref}
\usepackage{xcolor, pgf}
\usepackage{color, colortbl}
\usepackage[normalem]{ulem} 
\usepackage{placeins}
\usepackage{soul}
\usepackage[skins]{tcolorbox}
\usepackage{multirow}
\newcolumntype{C}[1]{>{\centering\arraybackslash}p{#1}}

\usepackage{algorithm}
\usepackage{algorithmicx}
\usepackage{algpseudocode}
\algnewcommand\algorithmicinput{\textbf{Input:}}
\algnewcommand\INPUT{\item[\algorithmicinput]}
\algnewcommand\algorithmicoutput{\textbf{Output:}}
\algnewcommand\OUTPUT{\item[\algorithmicoutput]}
\algnewcommand\algorithmicforeach{\textbf{for each}}
\algtext*{EndFor}

\algdef{S}[FOR]{ForEach}[1]{\algorithmicforeach\ #1\ \algorithmicdo}
\algrenewcommand{\alglinenumber}[1]{\color{red!80!blue}\footnotesize#1:}

\algnewcommand\Func[2]{\textcolor{green}{#1}\textcolor{green}{(#2)}}
\algnewcommand\Insert[2]{Insert {#1} to #2.}
\algnewcommand\Input[1]{\State \textbf{Input: } #1}
\algnewcommand\Output[1]{\State \textbf{Output: } #1}

\newlength\myboxwidth
\definecolor{gray}{rgb}{0.5,0.5,0.5}
\definecolor{green}{rgb}{0, 0.6, 0}
\definecolor{orange}{rgb}{1, 0.5, 0}
\definecolor{mahogany}{rgb}{0.75, 0.25, 0.0}
\definecolor{purple}{rgb}{0.6, 0, 0.6}
\definecolor{darkgreen}{rgb}{0, 0.3, 0}
\definecolor{orange}{rgb}{1, 0.5, 0.}
\definecolor{lightblue}{rgb}{0.52, 0.75,0.91}
\definecolor{softgreen}{rgb}{0.66,0.87,0.74}
\definecolor{softred}{rgb}{0.96,0.71,0.69}

\newcommand{\sgphl}[1]{#1}

\newcommand{\better}[1]{\textbf{#1}}

\newcommand{\ignore}[1]{}
\newcommand{\none}[1]{}
\newcommand{\com}[1]{}

\newcommand{\etal}{{\textit{et~al.}}}
\newcommand{\ie}{i.e.,}
\newcommand{\eg}{e.g.,}
\newcommand{\figname}{Figure}
\newcommand{\tabname}{Table}
\newcommand{\secname}{Section}

\newcommand{\eqname}{Eq.}

\DeclareMathOperator*{\argmin}{arg\,min}

\newcommand{\prjname}{StylePart}
\newcommand{\bm}[1]{\mathbf{#1}}

\begin{document}

\title[StylePart: Image-based Shape Part Manipulation]%
      {StylePart: Image-based Shape Part Manipulation}


\author[I-Chao Shen et al.]
{\parbox{\textwidth}{\centering I-Chao Shen$^{1}$  ~ Li-Wen Su$^{2}$ ~ Yu-Ting Wu$^{3}$ ~ Bing-Yu Chen$^{2}$
        }
        \\
{\parbox{\textwidth}{\centering $^1$ The University of Tokyo, Japan \\
$^2$ National Taiwan University, Taiwan \\
$^3$ National Taipei University, Taiwan
      }
}
}

\teaser{
 \includegraphics[width=1.0\linewidth]{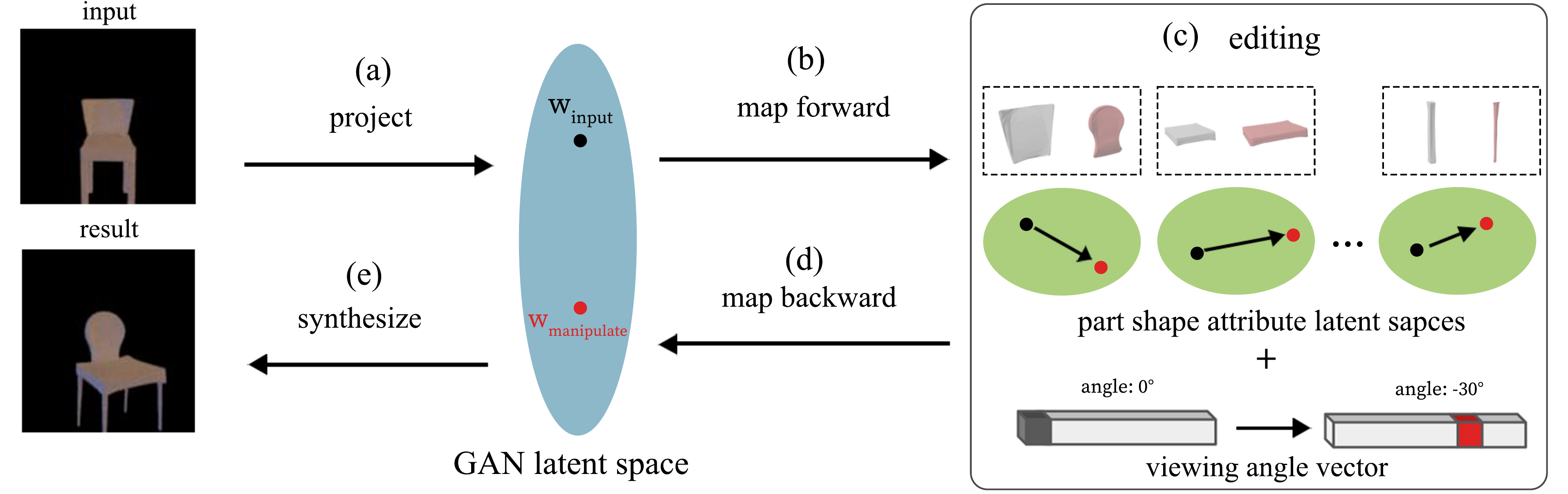}
 \centering
  \caption{
  \textbf{Overview of \prjname.}
  (a) We first project the input image into the GAN latent space, and (b) map the projected GAN latent code $\mathbf{w}_\text{input}$ to its corresponding shape attributes and viewing angle using a forward shape-consistent latent mapping function.
  (c) A user can directly manipulate the image shape at the part-level. Our method involves manipulation of the shape attributes and viewing angle vector. 
  Then, we obtain the manipulated GAN latent code $\mathbf{w}_\text{manipulate}$ by (d) mapping the manipulated attributes to the GAN latent space with a backward mapping function.
  Finally, we (e) synthesize the final edited image.
  }
\label{fig:teaser}
}

\maketitle
\begin{abstract}
Due to a lack of image-based ``part controllers'', shape manipulation of man-made shape images, such as resizing the backrest of a chair or replacing a cup handle is not intuitive.
To tackle this problem, we present \prjname, a framework that enables direct shape manipulation of an image by leveraging generative models of both images and 3D shapes.
Our key contribution is a shape-consistent latent mapping function that connects the image generative latent space and the 3D man-made shape attribute latent space.
Our method ``forwardly maps'' the image content to its corresponding 3D shape attributes, where the shape part can be easily manipulated.
The attribute codes of the manipulated 3D shape are then ``backwardly mapped'' to the image latent code to obtain the final manipulated image.
We demonstrate our approach through various manipulation tasks, including part replacement, part resizing, and viewpoint manipulation, and evaluate its effectiveness through extensive ablation studies.



\end{abstract}

\section{Introduction}
\label{sec:intro}

Manipulating 3D objects based on a single photo is a challenging task.
It involves various complex subtasks: \eg~extracting the target shape from the background; estimating the 3D pose, shape, and materials of the object; 
estimating the lighting conditions in the scene through  image observation; and extracting controllers to manipulate the estimated object properties.
Recently, several advanced deep learning-based methods have been developed to perform numerous subtasks, including object segmentation~\cite{He_2017_ICCV}; object pose~\cite{Song_2020_CVPR}, shape~\cite{Feng:SIGGRAPH:2021, meshrcnn,pix2surf_2020}, and material estimation~\cite{material2017}; and lighting estimation~\cite{Garon_2019_CVPR,Zhu_2021_CVPR}.

Moreover, generative adversarial networks (GANs) have opened up a new high-fidelity image generation paradigm. 
For example, because of its disentangled style space, StyleGAN~\cite{karras2019stylegan} can produce high-resolution facial images with unmatched photorealism and support stylized manipulation.
A user can transform the generated outputs by tweaking the latent code~\cite{shen2020interpreting,shen2021closedform,harkonen2020ganspace,gansteerability}.
A user can also edit a natural image by projecting it into the GAN image latent space, finding a latent code that reconstructs the input image and then modifying that code~\cite{zhu2016generative,chong2020ganui}.


Semantic 3D controllers are introduced for semantic parameter control over images.
For example, for generated~\cite{tewari2020stylerig} and natural~\cite{tewari2020pie,tewari2017mofa} human face images, the 3D morphable face model (3DMM) has been used to control facial shape, facial expression, head orientation and scene illumination.
However, the proposed methods apply only to portrait face images.

In this paper, we present \prjname, \sgphl{which investigates how to achieve part-based manipulation of man-made objects in a single image.}
The key insight of our method is to augment a 3D generative model representation that is controllable and with semantic information with the image latent space.
We propose a shape-consistent mapping function that connects the image generative latent space and latent space of the 3D man-made shape attribute.
The shape-consistent mapping function is composed of forward and backward mapping functions.
We ``forwardly map'' the input image from the image latent code space to the shape attribute space, where the shape can be easily manipulated.
The manipulated shape is re-mapped back to the image latent code space and synthesized using a pretrained StyleGAN.
We also propose a novel training strategy that guarantees the soundness of the mapped 3D shape structure.

\sgphl{
Note that we focus on the subtask of extracting controllers to manipulate the shape only, instead of proposing a full image manipulation system that tackle all subtasks altogether.
Thus we use images with a simple rendering style without complicated lighting conditions and material properties to demonstrate our results.
}
We evaluate our method through shape reconstruction tests, considering four man-made objectcategories (chair, cup, car, and guitar).
We also present several \textit{identity-preserving} manipulated results of three shape-part manipulation tasks: including part replacement, part resizing, and viewpoint manipulation.
\section{Related Work}
\label{sec:related}
\subsection{Image-based shape reconstruction and editing}
Three-dimensional modeling based on a single photo has been a challenge in the field of computer graphics and computer vision.
Several studies have investigated shape inference from multiple images~\cite{Paul:1996,SSS:2006} and single photos  for different shape representations, such as voxel~\cite{marrnet,NIPS2016_6206}, point cloud~\cite{su2017a}, mesh~\cite{meshrcnn,mao2020std,pix2surf_2020}, and simple primitives~\cite{chen20133}.
With advances in deep learning methods, the quality of reconstructed shapes and images has improved tremendously; however, they are usually not semantically controllable.
Chen~\etal~\cite{chen20133} proposed \textit{3-Sweep}, an interactive method for shape extraction and manipulation in a photo.
A user can create 3D primitives (cylinders and cuboids) using \textit{3-Sweep} and manipulate the photo content using extracted primitives.
Banerjee~\etal~\cite{OM3D2014} enable users to manually aligned a publicly available 3D model to guide the completion of geometry and light estimations.
Unlike previous methods, our method leverages recent advances in part-based generative shape representations~\cite{gao2019sdm,mo2019structurenet} to automatically infer shape attributes.
Moreover, the shape parts are more complex than simple cylinders and cuboids.

\subsection{GAN inversion and latent space manipulation}
GAN inversion is required to edit a real image through latent manipulation.
GAN inversion identifies the latent vector from which the generator can best replicate the input image. 
Inversion methods can typically be divided into optimization- and encoder-based methods.
Optimization-based methods which directly optimize the latent code using a single sample~\cite{abdal2019image2stylegan, abdal2020image2stylegan++, creswell2018inverting,huh2020transforming}, whereas encoder-based methods train an encoder over a large number
of samples~\cite{guan2020collaborative, tov2021designing,alaluf2021restyle}.
Some recent works have augmented the inversion process with additional semantic constraints~\cite{zhu2020indomain} and additional latent codes~\cite{gu2020image}.
Among these works, many have specifically considered StyleGAN inversion and investigated latent spaces with different disentanglement abilities, such as $\mathcal{W}$~\cite{karras2020analyzing}, $\mathcal{W}^{+}$~\cite{abdal2019image2stylegan,abdal2020image2stylegan++}, and style space ($\mathcal{S}$)~\cite{wu2021stylespace}.
In our work, we use StyleGAN-ADA~\cite{karras2020training} as the generator and we follow its original inversion process in the $\mathcal{W}$ space.
Moreover, we use additional shape attribute spaces to facilitate semantic disentanglement and support various image-based shape manipulation tasks.

Several works have examined semantic directions in the latent spaces of pretrained GANs. 
Full-supervision using semantic labels~\cite{Goetschalckx2019GANalyze,shen2020interpreting} and self-supervised approaches~\cite{jahanian2019steerability,plumerault2020controlling,spingarn2020gan} have been employed. 
Moreover, several recent studies have utilized unsupervised methods to obtain semantic directions\cite{wang2021geometry,harkonen2020ganspace,shen2021closedform}.
A series of works focused on real human facial editing~\cite{tewari2020stylerig,tewari2020pie} and non-photorealistic faces~\cite{wang2021cross}; they utilized a prior in the
form of a 3D morphable face model.
\section{Method}
\begin{figure*}[h!]
\centering
\includegraphics[width=\linewidth]{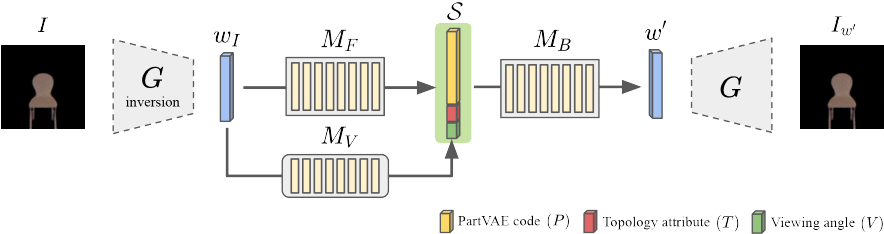}
   \caption{
   The network architecture of our cross-domain mapping framework.}
\label{fig:network}
\end{figure*}
We aim to enable users to directly edit the structure of a man-made shape in an image.
We first describe the background of the 3D shape representation method, which allows for tight semantic control of a 3D man-made shape.
We then describe a new neural network architecture that maps latent vectors between the image and 3D shape domains, and highlight the different shape part manipulation methods of the architecture.

\subsection{3D shape attribute representation}
We adopt a structured deformable mesh generative network (SDM-NET)~\cite{gao2019sdm} to represent the man-made shape attributes.
The network allows for tight semantic control of different shape parts.
Moreover, it generates a spatial arrangement of closed, deformable mesh parts, which represents the global part structure of a shape collection, \eg~chair, table, and airplane.
In SDM-NET, a complete man-made shape is generated using a two-level architecture comprising a part variational autoencoder (PartVAE) and structured parts VAE (SP-VAE).
PartVAE learns the deformation of a single part using
its Laplacian feature vector, extracted through the method established in~\cite{gao2019sparse}, and SP-VAE jointly learns the deformation of all parts of a shape and the structural relationship between them.
In this work, given an input shape $s$ of category $c$ with $n_{c}$ as the number of parts, we represent its shape attributes $\bm{S}=(\bm{P}, \bm{T})$ using both
\begin{itemize}
\item \textit{Geometric} attribute ($\bm{P}\in \mathbb{R}^{z\times n_{c}}$):  all the PartVAE latent codes of each shape, where $z=128$ is the dimension of the PartVAE latent code adopted in our work.
\item \textit{Topology} attribute ($\bm{T}\in \mathbb{R}^{2n_{c}+9}$): the representation vector $\mathbf{rv}$ in SDM-Net.
This vector represents the geometry and associated relationships of each part in $s$.
Please refer to \secname~3.1 in Gao~\etal~\shortcite{gao2019sdm}.
\end{itemize}

\subsection{Shape-consistent mapping framework}
At the core of our image-based shape part manipulation pipeline (shown in \figname~\ref{fig:network}) is a cross-domain mapping structure between the image latent space $\mathcal{W}$ and the shape attribute space $\mathcal{S}$.
Given an input image containing a man-made object ($I$), the image inversion method is first applied to optimize a GAN latent code ($w_{I}$) that can best reconstruct the image.
Then, the corresponding shape attributes ($\bm{S}$) are obtained using a forward shape-consistent latent mapping function ($M_{F}$).
The viewing angle of the input image is predicted using a pretrained viewing angle predictor ($M_{V}$).
From the mapped geometric attribute, topology attribute, and the predicted viewing angle, we can perform image-based shape editing tasks, such as part replacement, part deformation, and viewing angle manipulation by manipulating the attribute codes.
After manipulating the attribute codes, we use the backward mapping function ($M_{B}$) to map the shape attributes to an image latent code (${w}'$).
The final edited image can be synthesized as $I_{{w}'} = G({w}')$.
\subsubsection{Image inversion}
To obtain the latent code of the input image $I$, we optimize the following objective:
\begin{align} 
w_{I} = \argmin_{w} \mathcal{L}_{LPIPS}(I, G(w;\theta))+\lambda_{w}\mathcal{L}_{\mathbf{w}_\text{reg}}(w),
\label{gan-inversion-eq}
\end{align}
where $G$ is a pretrained StyleGAN-ADA~\cite{Karras2020ada} generator with weight $\theta$, $\mathcal{L}_{LPIPS}$ denotes the perceptual loss, and $\mathcal{L}_{\mathbf{w}_\text{reg}} = \|w\|^2_2$ denotes the latent code regularization loss.
We introduced $\mathcal{L}_{\mathbf{w}_\text{reg}}$ above because we observed that there are multiple latent codes that can synthesize the same image. 
By introducing $\mathbf{w}_\text{reg}$, we obtain a unique latent code for each input image.
\subsubsection{From $\mathcal{W}$ to $\mathcal{S}$}
Given a GAN latent code ${w_I}$, which is inverted from an input image $I$, we obtain the corresponding shape attribute code $\bm{S}=(\bm{P}, \bm{T})$ using the forward mapping network $(\bm{P}, \bm{T}) = M_F(\bm{w_I})$, where the 3D shape generated by $(\bm{P}, \bm{T})$ best fits the target man-made shape encoded in the image latent code $\bm{w_I}$.

\subsubsection{From $\mathcal{S}$ to $\mathcal{W}$}
Given the shape attribute $\bm{S}$ and one-hot vector of a viewing angle $\bm{v}$, the backward mapping function predicts a GAN latent code $\mathbf{w}' = M_{B}(\bm{S}, \bm{v})$, where the synthesized image $I_{{w}'} = G({w}')$ best matches the image containing the target shape described in $\bm{S}$ with viewing angle $\bm{v}$.

Our mapping functions ($M_{F}$ and $M_{B}$) are realized using two eight-layer MLP networks (see the supplemental material for the detailed architecture).

\subsection{Training strategy and loss function}
\label{sec:train_and_loss}
\subsubsection{Data preparation}
To train our shape-consistent mapping function, both the image-based latent code and the shape attribute code for each man-made object shape are required.
In this paper, we use synthetic datasets of four categories (chair, guitar, car, and cup). 
A dataset contains $N^M$ shapes, formed by interchanging the $M$ parts of $N$ shapes.
We render these shapes according to different viewing angles to obtain the paired shapes and images. 
We sample the viewing angles at $30^{\circ}$ intervals on the yaw axis.
For the chair and guitar category, $N=15$ and $M=3$; for the cup and car categories, $N=50$ and $M=2$.
There are $3,375$ shapes and $40,500$ images for the chair and guitar categories, and $2,500$ shapes and $30,000$ images for the car and cup category.
We split the synthetic datasets for each category into $80\%$ training data and $20\%$ testing data
For shape $k$, we first prepare the shape attribute code $\bm{S}_k$.
Each part is represented by a feature matrix $f\in\mathbb{R}^{9\times V}$ that describes the per-vertex deformation of a template cube with $V=3,752$ vertices.
Let $n_c$ denote the number of the parts for category $c$, and $z_{part}=128$ is the dimension of a PartVAE latent code. 
We embed the feature matrix of part $i$ ($f_i$) into a pretrained PartVAE and obtain $\bm{P}_i = Enc_{P_i}(f_i)$.
We prepare the same topology vector $\bm{T} \in R^{2n+73}$ for shapes in the same category.
Finally, for each shape, we concatenate the PartVAE codes of all parts and the topology vector into the final shape attribute vector $\bm{S}=(\bm{P}_0, \bm{P}_1, ..., \bm{P}_{n-1}, \bm{T})$.

Next, we prepare the image-based latent code for shape $k$.
We render all shapes from 12 viewing angles and use these images to train a StyleGAN2 through adaptive discriminator augmentation \cite{Karras2020ada}, which depends on the image's viewing angle.
We project the rendered image containing shape $k$ into the pretrained conditional StyleGAN latent space to obtain the corresponding latent code $\bm{w}_k \in R^{512}$.
We collect all $(\bm{w}_k, \bm{S})$ pairs for training both $M_{F}$ and $M_{B}$.
\subsubsection{Training for $M_F$}
\begin{figure}
\centering
\includegraphics[width=\linewidth]{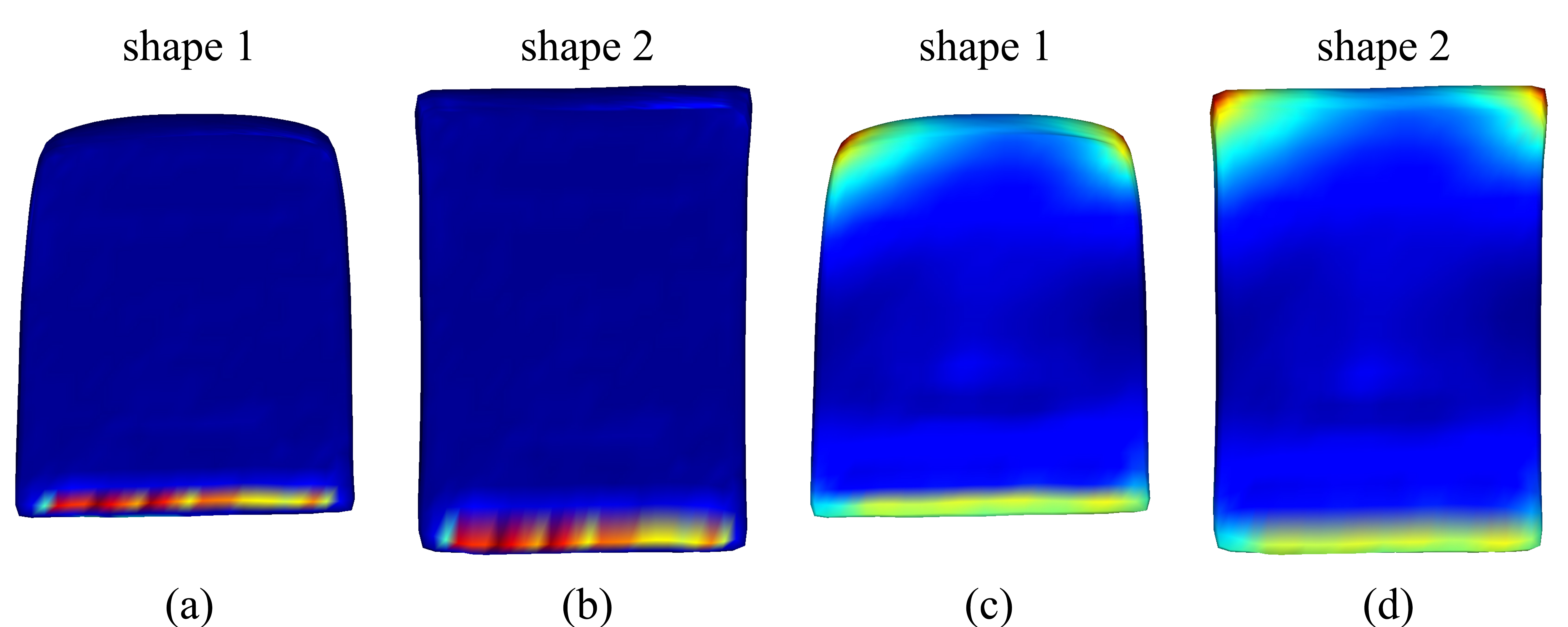}
   \caption{
   Visualization of Laplacian feature differences (a, b) and vertex coordinate differences (c, d) between shape 1 and 2.
   In contrast to Laplacian feature differences, vertex coordinate differences capture the deformations. 
   }
\label{fig:lap_vs_vertex}
\end{figure}

We train the forward mapping function $M_{F}$ using a two-step process comprising \textit{Laplacian feature reconstruction training} and \textit{size finetuning}. 
In the Laplacian feature reconstruction training step, we use the following loss function:
\begin{align}
\mathcal{L}_{M_{F}} = \mathcal{L}_{\bm P_\text{recon}} + \mathcal{L}_{\bm T},
\end{align}
where $\mathcal{L}_{\bm P_\text{recon}}$ and $\mathcal{L}_{\bm T}$ are defined as follows:
\begin{align}
    \mathcal{L}_{\bm P_\text{recon}} &= \sum_{i=1}^{{N}} ||{Dec}_{P_i}(\bm{P'}) - {Dec}_{P_i}(\bm{P})||_2^2 \\
    \mathcal{L}_{\bm T} &= \sum_{i=1}^{{N}} ||\bm{T'} - \bm{T}||_2^2
\end{align}
Here, $(\bm{P'}, \bm{T'})$ are the shape attrbutes predicted by $M_{F}$, $Dec_{P_i}(\cdot)$ denotes the pretrained PartVAE decoder of the $i$-th part, and ${N}$ denotes the number of parts.
In the second step, we replace the reconstructed feature vectors from the Laplacian feature to the vertex coordinates. 
As shown in \figname~\ref{fig:lap_vs_vertex}, we observed the Laplacian feature vectors are sensitive to local differences but insensitive to global shape differences.
The loss function in this step can be written as:
\begin{align}
\hat{\mathcal{L}}_{M_{F}} &= \mathcal{L}_{\bm V_\text{recon}} + \mathcal{L}_{\bm T} \\
\mathcal{L}_{\bm V_\text{recon}} 
&= \sum_{i=1}^{{N}} ||\mathcal{T}({Dec}_{P_i}(\bm{P'})) - \mathcal{T}({Dec}_{P_i}(\bm{P}))||_2
\end{align}
where $\mathcal{T}$ transforms a vertex Laplacian feature vector into its vertex coordinates according to the steps described in \cite{gao2019sparse}.
\subsubsection{Training for $M_B$}
Our backward mapping function $M_{B}$ minimizes the following loss:
\begin{equation}
    \mathcal{L_{\textit M_{\textit B}}} = \mathcal{L_{\mathbf{w}_\text{recon}}} + \lambda_1 \mathcal{L_{\mathbf{w}_\text{reg}}}
\end{equation}
where $\lambda_1$ is the weight of the $l_2$ norm regularization term of the image latent code $w'$ (the same weight used in image inversion in \eqname~\ref{gan-inversion-eq}), and $\mathcal{L_{\mathbf{w}_\text{recon}}}$ is defined as:
\begin{equation}
    \mathcal{L_{\mathbf{w}_\text{recon}}} = \mathcal{L}_{LPIPS}(G({w}'), G({w}))
\end{equation}
where $G$ is the pretrained Style-ADA generator, $w'$ is the mapped image latent code of $(\bm{P}, \bm{T})$, and $\mathcal{L}_{LPIPS}(\cdot)$ is the perceptual loss function.

\subsubsection{Finetuning for $M_F$ and $M_B$}
\begin{figure}
\centering
\includegraphics[width=1.0\linewidth]{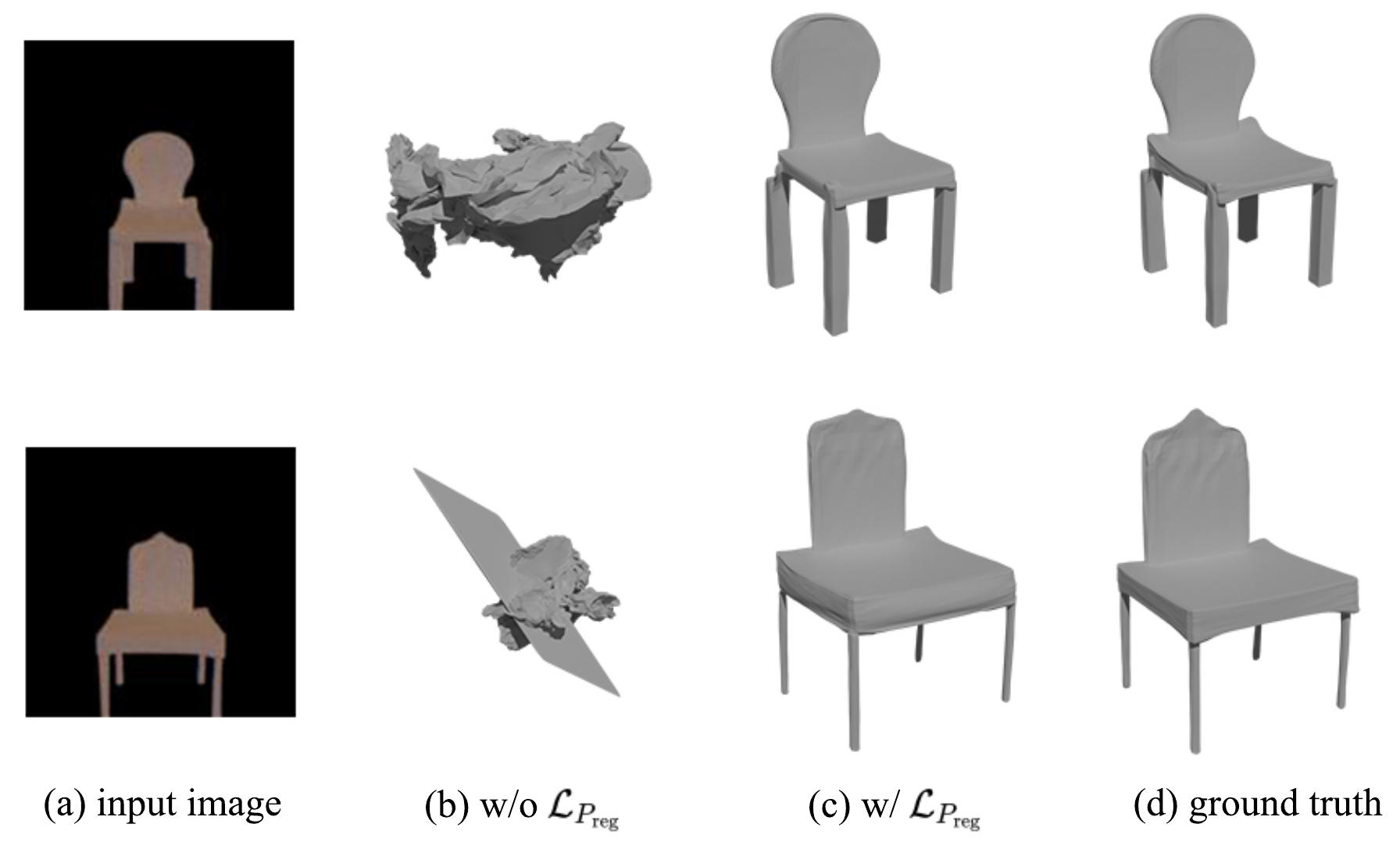}
   \caption{
   (a) Input image $I$; (b) is the mapped shape of $w_I$ without $\mathcal{L}_{\bm P_\text{reg}}$; (c) is the mapped shape of $w_I$ with $\mathcal{L}_{\bm P_\text{reg}}$; (d) is the target shape rendered in the input image.
   }
\label{fig:ft_s_reg}
\end{figure}

After training $M_{F}$ and $M_{B}$ separately, we finetune them together using the following loss function:
\begin{align}
    \mathcal{L_{\textit{finetune}}} = 
    \mathcal{L}_{\bm T} + \mathcal{L_{\mathbf{w}_\text{recon}}}  + \lambda_1 \mathcal{L_{\mathbf{w}_\text{reg}}} + \lambda_2 \mathcal{L}_{\bm P_\text{reg}}
\end{align}
where $\lambda_2$ is the weight of a shape attribute regularization term which can be written as:
\begin{align}
    \mathcal{L}_{\bm P_\text{reg}} =
    ||\textit{M}_{\textit F}(\mathit{w}) - \textit{M}_{\textit{F}_\text{freeze}}(\mathit{w})||_2
\end{align}
Here, we introduce a Laplacian feature reconstruction loss with a shape attribute regularization term $\mathcal{L}_{\bm P_\text{reg}}$; the loss function differs from the loss functions used to train $M_F$ and $M_B$.
In~\figname~\ref{fig:ft_s_reg}, we show the reconstructed shapes based on PartVAE latent codes predicted with or without $\mathcal{L}_{\bm P_\text{reg}}$.
The shape attribute regularization term prevents substantial deviation of the PartVAE latent codes from reasonable 3D shapes.

\section{Shape part manipulation}
We propose three types of manipulations: part replacement, part deformation, and viewing angle manipulation. 
\subsection{Part replacement}
\begin{figure}[t!]
\centering
\includegraphics[width=\linewidth]{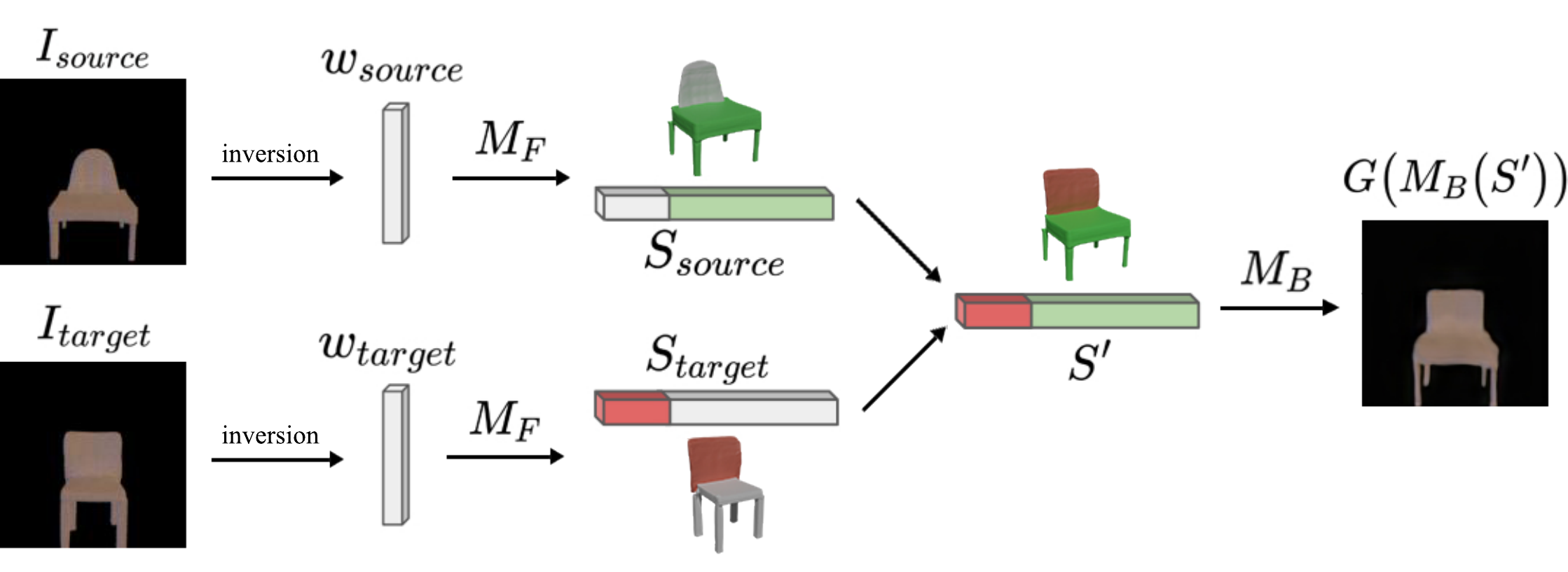}
   \caption{
   Part replacement editing process baed on our pipeline.
   }
\label{fig:replacement}
\end{figure}

The first manipulation task is to replace parts of the input shape.
Given the source image $I_{\text{source}}$, the user aims to replace one part (\eg~chair back) in $I_{\text{source}}$ with the corresponding part in the target image $I_{\text{target}}$.
The part replacement procedure of our mapping framework is illustrated in \figname~\ref{fig:replacement}. 
First, we obtain the image latent codes ($w_{\text{source}}$ and $w_{\text{target}}$) through image inversion and then we use the pretrained $M_F$ to obtain the shape attributes $\bm{S}_{\text{source}}$ and $\bm{S}_{\text{target}}$ of the shape in both input images. 
A user can select which part (\eg~back, seat, or leg) of the shape in $I_{\text{source}}$ he/she wants to replace.
The corresponding shape attribute representing the selected part is then replaced with the target shape attribute $\bm{S}'_{\text{source}}=\{\bm{P}_0^{\text{source}}, \bm{P}_1^{\text{target}}, ... \bm{P}_{n-1}^{\text{source}}\}$. 
Finally, we synthesize the edited image ${I}'= G(M_B({\bm{S}}', \bm{V}))$, where $\bm{V}$ is the original viewing angle vector. 
The new shape in the manipulated image will contain the part selected from $I_{\text{target}}$, while the non-selected parts will remain as close as possible to the original parts in $I_{\text{source}}$.

\subsection{Part resizing}
\label{sec:part_resize}
The second manipulation task is part resizing.
A user directly resizes a selected part in the input image.
We first invert an input image $I$ to obtain its GAN latent code $w_I$ and map $w_I$ to the shape attribute $\bm{S}$ by $M_F$.
The user can resize the selected part by following a certain trajectory in the latent space and obtain the resulted image $I_{\text{resize}}$:
\begin{align}
I_{resize} = G(w_{I}+\mathcal{F}_{r}(r^{\bm{P}})),
\end{align}
where ${r}^{\bm{P}}$ is the trajectory of the PartVAE latent code that represents the desired resizing result, and $\mathcal{F}_{r}$ is a trajectory finetuner function that refines a PartVAE code trajectory into a GAN latent code trajectory.
\subsubsection{Resize trajectory}
Trajectories that fit the desired resize manipulation in $\mathcal{S}$ and $\mathcal{W}$ are obtained using the following procedure.
Given a shape $s_i$ in our dataset, we obtain its geometric attribute by $\bm{P}_\text{i}={\text{Enc}}_{P}(\mathcal{T}^{-1}(s_i))$, where ${\text{Enc}}_{P}$ is a pretrained PartVAE encoder.
Then we apply a specific resizing operation to $s_i$ to obtain the resized shape $\hat{s}_i = \mathcal{R}(s_i)$ and its geometric attribute $\hat{\bm{P}}_i={\text{Enc}}_{P}(\mathcal{T}^{-1}(\hat{s}_i))$.
Thus, we obtain a PartVAE latent trajectory for this resize manipulation of $s_i$ by $r_i^{\bm{P}}=\hat{\bm{P}}_i-\bm{P}_i$. 
We apply the same resize manipulation to all shapes in our dataset and average all PartVAE latent trajectories to obtain a general trajectory $r^{\bm{P}}=(1/N)\sum_{N}(r_i^{\bm{P}})$ representing the resizing manipulation.
By applying $r^{\bm{P}}$, we observe that the resized images often lose some details, thus impairing shape identify (\figname~\ref{fig:resize}).
For better shape identity after part resizing, we introduce a \textit{GAN space trajectory finetuner} implemented using a four-layer MLP.
The main function of this finetuner is to transform the trajectories from $\mathcal{S}$ to $\mathcal{W}$.
The finetuner inputs are the GAN latent code of input image $w_I$ and a PartVAE latent trajectory of the target part, while the output is a trajectory in $\mathcal{W}$.
We train this trajectory finetuner using paired trajectory data $(r^{\bm{P}}, r_i^{\bm{W}})$ and by minimizing the following loss function:
\begin{align}
\mathcal{L}_r=\|\mathcal{F}_{r}(w_I, r^{\bm{P}})- r_i^{\bm{W}} \|^2_2.
\end{align}
To collect the paired trajectory data, we first add $r^{\bm{P}}$ to the PartVAE latent codes of all training shapes and obtain the rendered image latent codes (\ie~$\hat{w}_i$) by optimizing Eq.\ref{gan-inversion-eq}.
For shape $i$, we obtain the GAN latent space trajectory $r_i^{\bm{W}}$ corresponding to $r^{\bm{P}}$ by $r_i^{\bm{W}} = \hat{w}_i - w_I$.

We use the shapes in the datasets mentioned in \secname~\ref{sec:implelmentation} as the original shapes. 
For each part, we resize shapes by adding three weights (-0.5$r^{\bm{P}}$, +0.5$r^{\bm{P}}$, +1.0$r^{\bm{P}}$) to the specific shape attribute trajectory.
There are $10,125$ ($3,375\times 3$) shapes for each part of chair and guitar, and 
$30,375$ ($10,125 \times 3$ parts) shapes in total. 
There are $7,500$ ($2,500 \times 3$) shapes for each part of car and cup, and a total of $21,500$ ($7,500\times 3$ parts) and $15,000$ ($7,500. \times 2$ parts) shapes, respecitvely.

\subsection{Viewing angle manipulation}
The third editing task is viewing angle manipulation. 
To achieve this, we first train a viewing angle prediction network ($M_{V}$) to predict the viewing angle of an image $I$ from its GAN latent code $w_{I}$, \ie~$\bm{V}=M_{V}(w_{I})$.
The viewing angle prediction network is an eight-layer MLP trained with the cross-entropy loss. 
For each category, we use the rendered images described in \secname~\ref{sec:train_and_loss} and collect $36,000$ paired training data $(w, v)$ from $3,000$ shapes in 12 different viewing angles.

To manipulate the viewing angle of input image $I$, we obtain the shape attribute $\bm{S}$ of the shape in the input image using $M_F$.
We obtain the edited image with the manipulated viewing angle vector: ${I}'=G(M_B((\bm{S}, \bm{V}')))$.
\section{Experiment and Results}
\subsection{Implementation details}
\label{sec:implelmentation}
For each man-made object category, we trained the StyleGAN-ADA generator using a batch size of $64$ and learning rate of $0.0025$.
Both of our forward and backward mapping networks were trained using the Adam~\cite{kingma2014adam} optimizer for $3,000$ epochs and a batch size $64$.
All networks were trained using a Tesla V100 GPU.
We implemented our pipeline using PyTorch~\cite{NEURIPS2019_9015}. 
The forward mapping network were trained for $0.5$-$3$ days, depending on the number of part labels.
The backward mapping network were trained for $20$ hours, and the forward and backward mapping networks were finetuned together for $10$ hours.

\begin{figure}
\centering
\includegraphics[width=\linewidth]{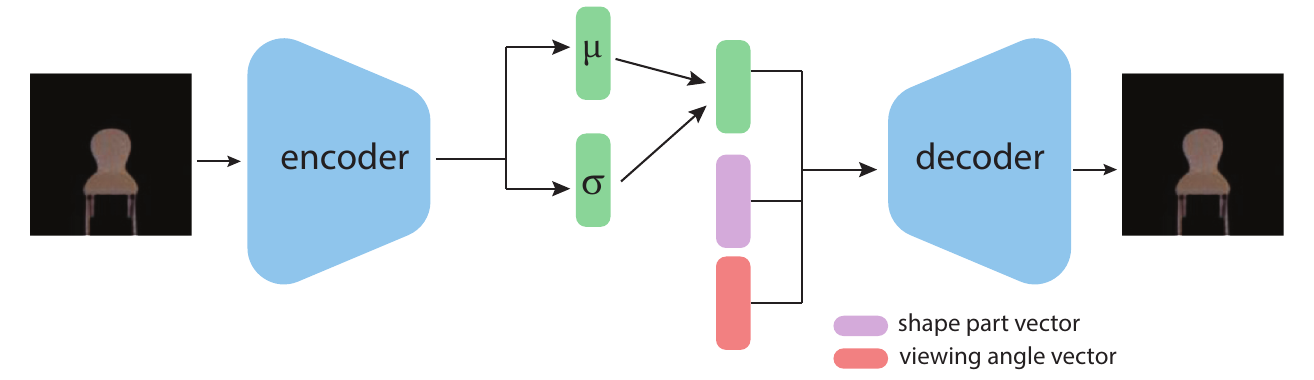}
   \caption{
The architecture of the baseline cVAE method.
   }
\label{fig:baseline_arch}
\end{figure}
\begin{description}[style=unboxed,leftmargin=0cm]
\item[Baseline disentangled network] 
We use a conditional variational autoencoder (cVAE) as the baseline disentangled method and compare our results with this baseline on image shape reconstruction and the results of viewing angle manipulation quantitatively. 
For each shape, we include three factors: part identity, part size, and viewing angle.
All these factors are modeled as discrete variables and represented as one-hot vectors during training and testing.
\end{description}

\subsection{Image shape reconstruction}
The key contribution of the proposed framework are the mapping functions between the GAN latent space and the shape attribute latent space.
We qualitatively and quantitatively evaluated these mapping functions through image shape reconstruction.
\subsubsection{With/Without size finetuning for $M_F$}
\begin{figure}
\centering
\includegraphics[width=1.0\linewidth]{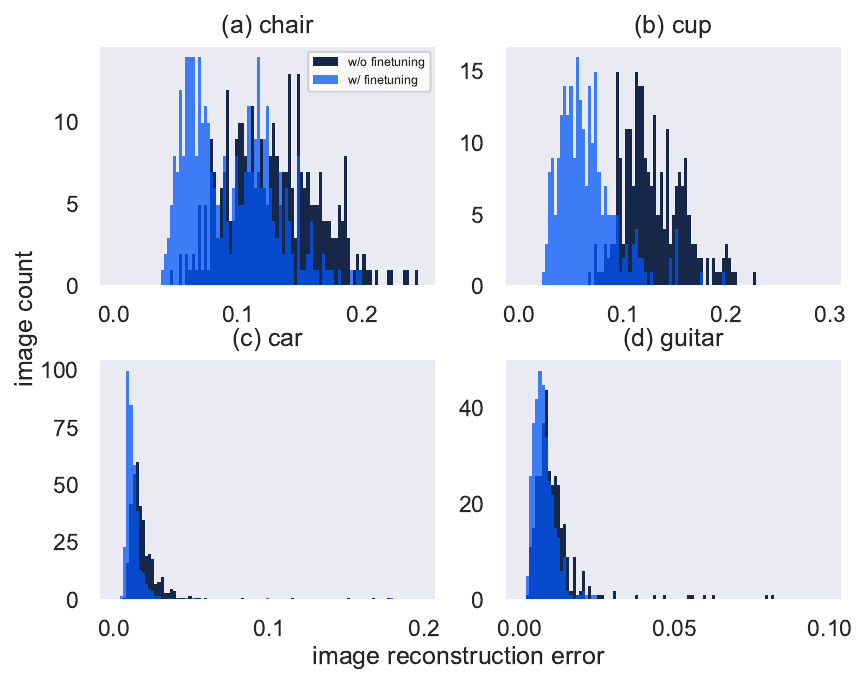}
  \caption{
  The perceptual loss distributions of the image shape reconstruction  with and without size finetuning.}
\label{fig:size_finetune_vgg}
\end{figure}

As described in \secname~\ref{sec:train_and_loss}, after the \textit{Laplacian feature reconstruction training}, we performed an additional \textit{size finetuning} for $M_F$ with vertex coordinate loss for a better reconstruction.
To test the effectiveness of the \textit{size finetuning}, we evaluated the reconstruction quality of the front view of the testing images described in \secname~\ref{sec:train_and_loss}.
We sampled $4,000$ points on each shape and calculated the bidirectional Chamfer distance as our shape reconstruction error metric, and we used perceptual loss as the image reconstruction error metric.
Both the image reconstruction and the shape reconstruction errors were lower when the network was trained with \textit{size finetuning} (\tabname~\ref{tab:w/wo_size_finetuneloss}(a)).
The image reconstruction perceptual loss distribution (\figname~\ref{fig:size_finetune_vgg}) also showed that for the chair and cup categories, the error distributions were significantly lower when the network was trained with \textit{size finetuning}.

\begin{table}
\footnotesize
\centering
\label{tab:recon_error}
    \begin{tabular}{l C{2.5mm}C{2.5mm}C{2.5mm} C{2.5mm}C{2.5mm}C{2.5mm} C{2.5mm}C{2.5mm}C{2.5mm}C{2.5mm}C{2.5mm}C{2.5mm}}
    \toprule
\multirow{2}{*}{}
        &   \multicolumn{3}{c}{\textbf{Chair}}
        &   \multicolumn{3}{c}{\textbf{Car}} 
        &   \multicolumn{3}{c}{\textbf{Cup}} 
        &   \multicolumn{3}{c}{\textbf{Guitar}} \\
        \cmidrule(lr){2-4}\cmidrule(lr){5-7} \cmidrule(lr){8-10} \cmidrule(lr){11-13}
    &   w/ &   w/o  & $\mathcal{B}$   &  w &   w/o & $\mathcal{B}$  &   w &   w/o  & $\mathcal{B}$ &  w &   w/o & $\mathcal{B}$ \\
    \midrule
$E_{s}\downarrow$ &\better{1.28} & 2.63 & -- & \better{1.45} & 3.07 & -- & \better{0.99} & 1.20 & -- & \better{0.40} & 0.74 & -- \\
$E_{i}\downarrow$ & \better{0.09} & 0.12 & 0.14 & \better{0.01} & 0.02 & 0.06 & \better{0.07} & 0.12 & 0.21 & \better{0.01} & 0.08 & 0.13\\
\bottomrule
\end{tabular}
\caption{
Comparison of the mean shape reconstruction errors $(E_{s})$ and mean image reconstruction errors $(E_{i})$ of results obtained from with and without size finetuning, and the baseline conditional VAE (cVAE) method $\mathcal{B}$.
We did not report the mean shape reconstruction error fo the baseline method because it did not reconstruct a explicit shape.
}
\label{tab:w/wo_size_finetuneloss}
\end{table}

\subsubsection{Finetuning for $M_F$ and $M_B$}
After separately training $M_F$ and $M_B$ , we performed an end-to-end finetuning to improve the reconstruction results.
We compared the full and non-finetuned versions of our method.
The full version included the shape attribute regularizer $\mathcal{L}_{\bm P_\text{reg}}$ to prevent the shape from collapsing.
We tested on about $300$ images in each category and obtained the mean perceptual losses of finetuned networks and non-finetuned networks. 
Through this step, the image reconstruction errors for chair, car, cup, and guitar categories were reduced by $28\%$, $37\%$, $27\%$, and $32\%$ respectively.

\subsection{Shape part manipulation results}
\subsubsection{Part replacement results}
We randomly picked $12$ different chair images and exchanged their parts.
We applied several replacement patterns to all possible combinations of the picked 12 images and produced edited images for each category. 
We show some replacement results in \figname~\ref{fig:replace_result}.
Detailed results are available in the supplementary material.

\begin{figure*}[h!]
\centering
\includegraphics[width=\linewidth]{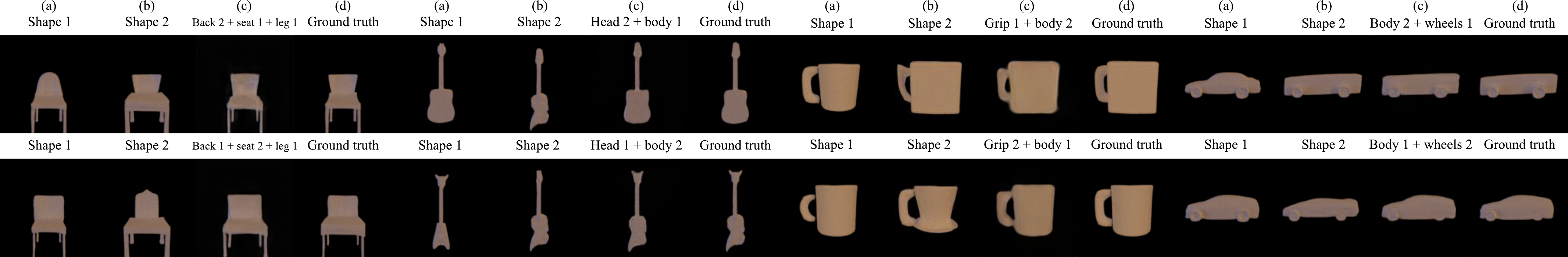}
   \caption{
   Results of part replacement. 
   The images were produced from the attributes of shape 1 and 2.}
\label{fig:replace_result}
\end{figure*}

\subsubsection{Part resizing results}
\begin{figure}
\centering
\includegraphics[width=\linewidth]{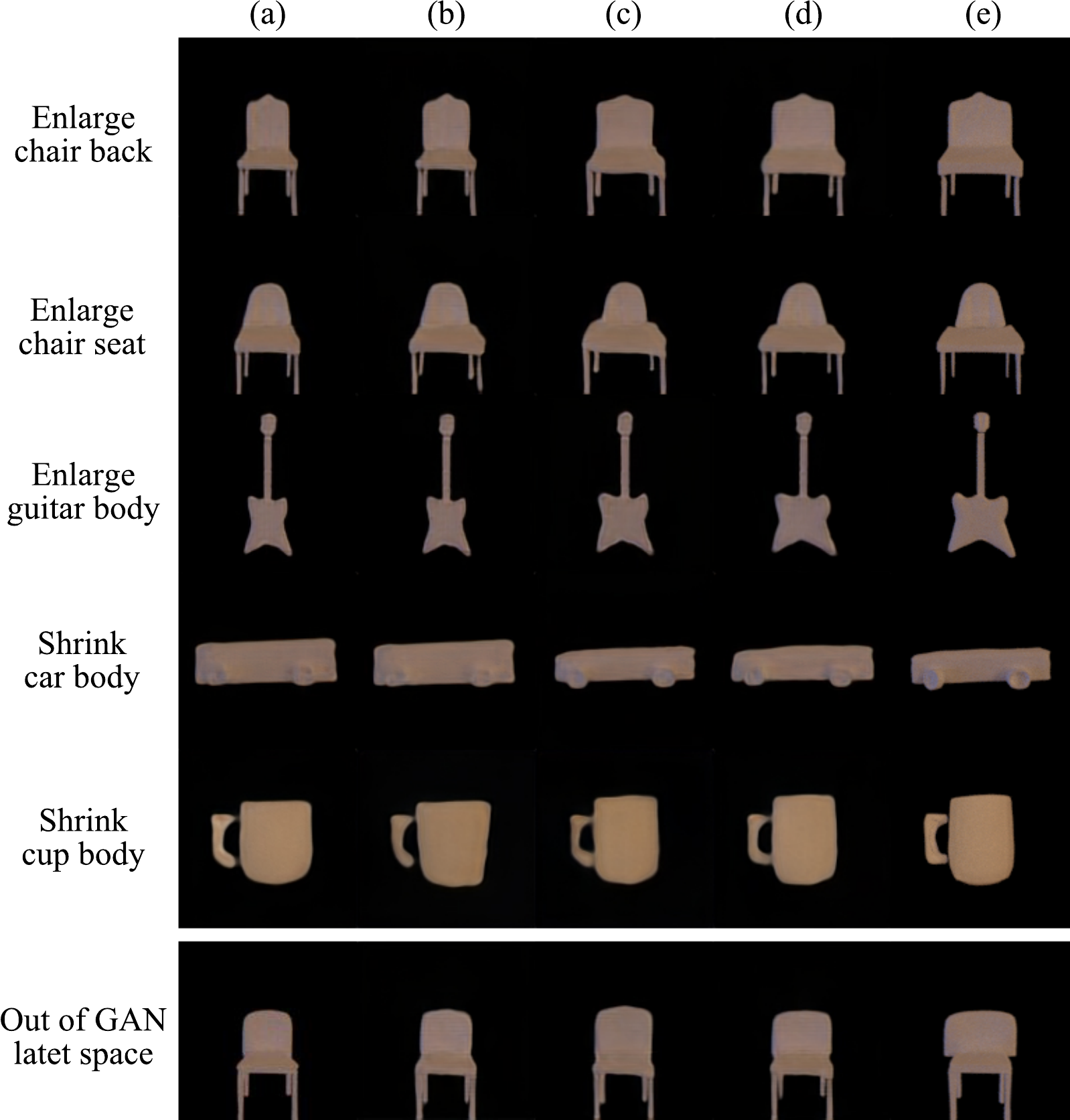}
   \caption{
    Part resizing results for different categories. (a) is the input image, (b) resized image directly obtained using $P+r^{\bm{P}}$; (c) resized image obtained using GAN space trajectory finetuner; (d) is the inverted result of ground truth, and (e) is the rendered ground truth.
   }
\label{fig:resize}
\end{figure}

We trained separate GAN latent trajectory finetuners for each part using the resizing dataset discussed in \secname~\ref{sec:part_resize}.
We compared the test results obtained by two methods: using shape attribute trajectory ($r^{\bm{P}}$) and the GAN space trajectory finetuner ($\mathcal{F}_{r}$). 
The comparison results and mean perceptual losses are shown in \figname~\ref{fig:resize} and \tabname~\ref{tab:resize_loss}, respectively. 
The manipulation result obtained from $\mathcal{F}_{r}$ showed less perceptual loss and matched the desired resize manipulation better than the results obtained from $r^{\bm{P}}$.
Moreover, we applied multiple trajectory finetuners trained on different parts of the same image to simultaneously resize multiple parts.
Regarding the failure case (last row in \figname~\ref{fig:resize}), it fails because the target resized image was outside the GAN latent space, and the finetuner could not find a good trajectory to fit the target image.

\begin{table}
\footnotesize
\centering
\label{tab:recon_error}
    \begin{tabular}{C{1mm} C{3.5mm}C{3.5mm}C{3.5mm} C{4mm}C{4mm} C{4mm}C{4mm} C{3.5mm}C{3.5mm}C{3.5mm}}
    \toprule
    \multirow{3}{*}{}
        &   \multicolumn{3}{c}{\textbf{Chair}}
        &   \multicolumn{2}{c}{\textbf{Car}} 
        &   \multicolumn{2}{c}{\textbf{Cup}} 
        &   \multicolumn{3}{c}{\textbf{Guitar}} \\
        \cmidrule(lr){2-4} \cmidrule(lr){5-6} \cmidrule(lr){7-8} \cmidrule(lr){9-11}
        & back  &  seat &  legs &  body   &   wheel &  body  &   grip &   head & neck & body  \\
    \midrule
 $r^{\bm{P}}$ & 8.22 & 11.95 & 5.60 & 13.70 & 7.05 & 14.55 & 12.67 & 1.62 & 2.85 & 4.62\\
$\mathcal{F}_{r}$ & \better{6.18} & \better{7.73} & \better{5.33} & \better{6.34} & \better{4.42} & \better{6.88} & \better{5.85} & \better{0.96} & \better{1.58} & \better{2.08}\\
\bottomrule
\end{tabular}
\caption{
Quantitative results for the part resizing manipulation.
Comparison of mean perceptual losses of results obtained using shape attribute trajectory ($r^{\bm{P}}$) and the GAN space trajectory finetuner ($F_{r}$).
}
\label{tab:resize_loss}
\end{table}


\subsubsection{Viewing angle manipulation results}
\label{sec:view_manipulation}
\begin{figure*}[h!]
\centering
\includegraphics[width=\linewidth]{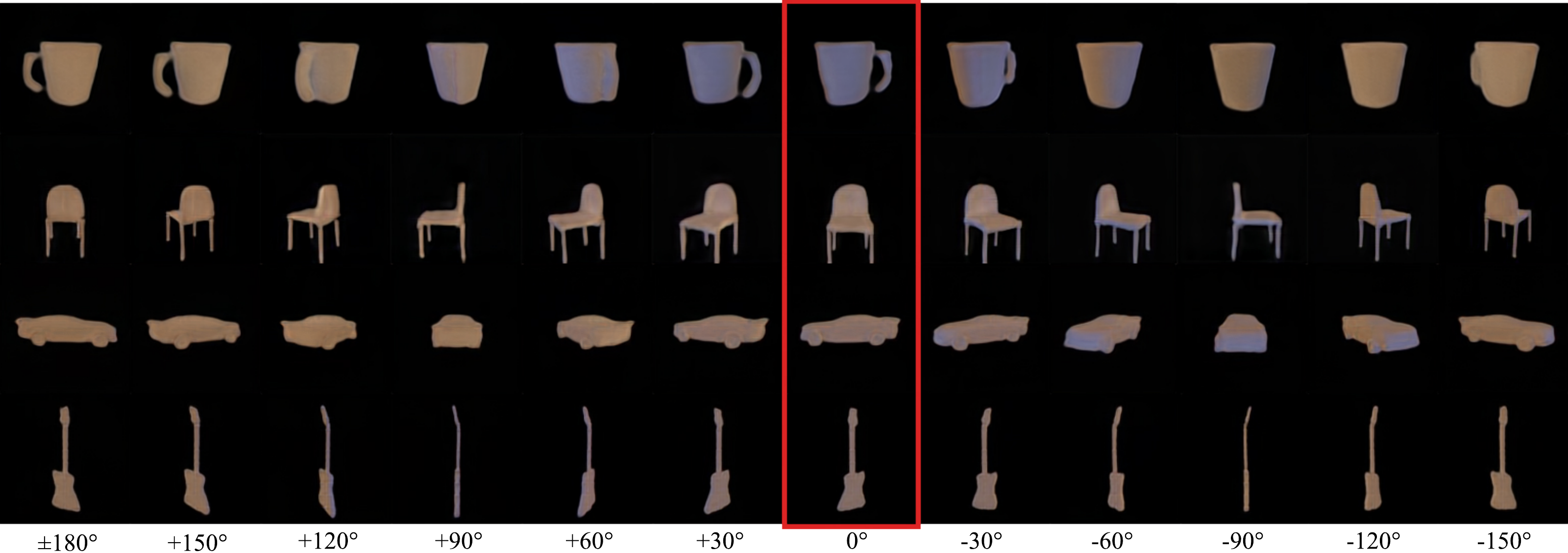}
   \caption{
   Viewing angle manipulation results. Our method can synthesize images of the shape from different viewing angles.
   }
\label{fig:multiview}
\end{figure*}

\begin{figure}
\centering
\includegraphics[width=\linewidth]{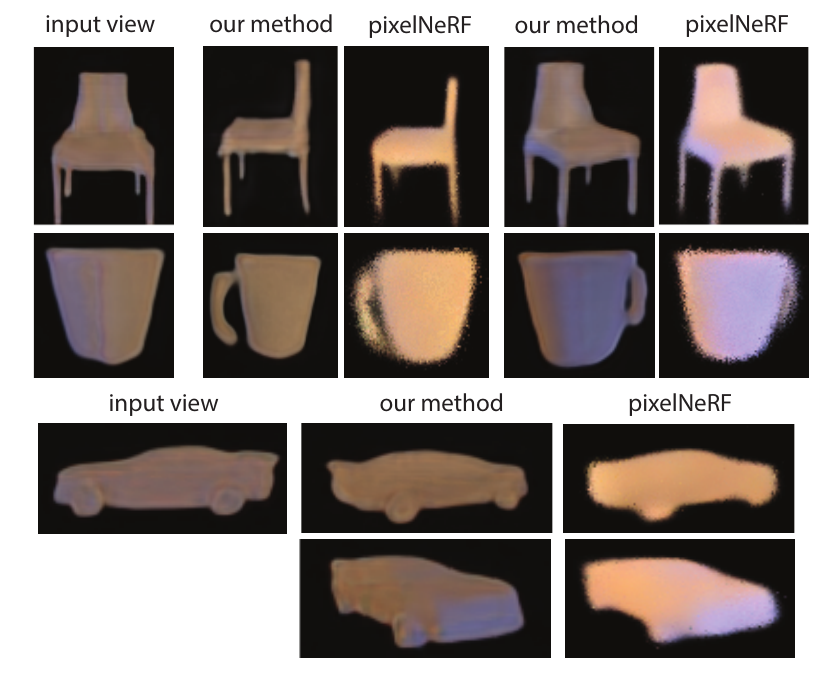}
   \caption{
   \textbf{Comparison with pixelNeRF~\cite{Yu_2021_CVPR}.}
   Our method synthesized shapes with more details while pixelNeRF only synthesize blurry shapes using the same input view.
   }
\label{fig:pixelnerf_comp}
\end{figure}

\figname~\ref{fig:multiview} shows the viewing angle manipulation results.
By manipulating the one-hot vector of the viewing angle of a given image, our method can produce the images from different viewing angles.
\sgphl{
We tested the images described in \mbox{\secname~\ref{sec:train_and_loss}} against the baseline method.
For each shape, each method synthesizes images from 12 different viewing angles as described in \mbox{\secname~\ref{sec:view_manipulation}}.
In \mbox{\tabname~\ref{tab:view_quan}}, we show mean perceptual losses across different viewing angles.
The results of our method obtain lower perceptual losses compared to the results of the baseline method (cVAE) and pixelNeRF~\mbox{\cite{Yu_2021_CVPR}}.
We trained a 1-view pixelNeRF using our training data.
This result suggested that the explicit shape attribute latent space helps to build up better geometries compared to random latent vectors and implicit geometries learned by radiance field-based method~\mbox{\cite{Yu_2021_CVPR}}.
}
The shape identities of the manipulated results (\figname~\ref{fig:multiview}) were well maintained.
\sgphl{
In \mbox{\figname~\ref{fig:pixelnerf_comp}}, we showed the visual comparisons of our results and the results of pixelNeRF~\mbox{\cite{Yu_2021_CVPR}}.
The results of pixelNeRF often fail to reconstruct the geometry details of shape parts, thus the shape identifies are not maintained.
}
\begin{table}[ht]
\centering
\begin{tabular}[t]{ccccc}
\toprule
&\textbf{Chair}$\downarrow$&\textbf{Car}$\downarrow$&\textbf{Cup}$\downarrow$&\textbf{Guitar}$\downarrow$\\
\midrule
Our&\better{0.0022}&\better{0.0075}&\better{0.0029}&\better{0.0043}\\
$\mathcal{B}$&0.0416&0.05&0.07&0.02\\
pixelNeRF~\cite{Yu_2021_CVPR}&&&&\\
\bottomrule
\end{tabular}
\caption{
Viewing anlge manipulation quantitative evaluation.
}
\label{tab:view_quan}
\end{table}%

\begin{figure}
\centering
\includegraphics[width=\linewidth]{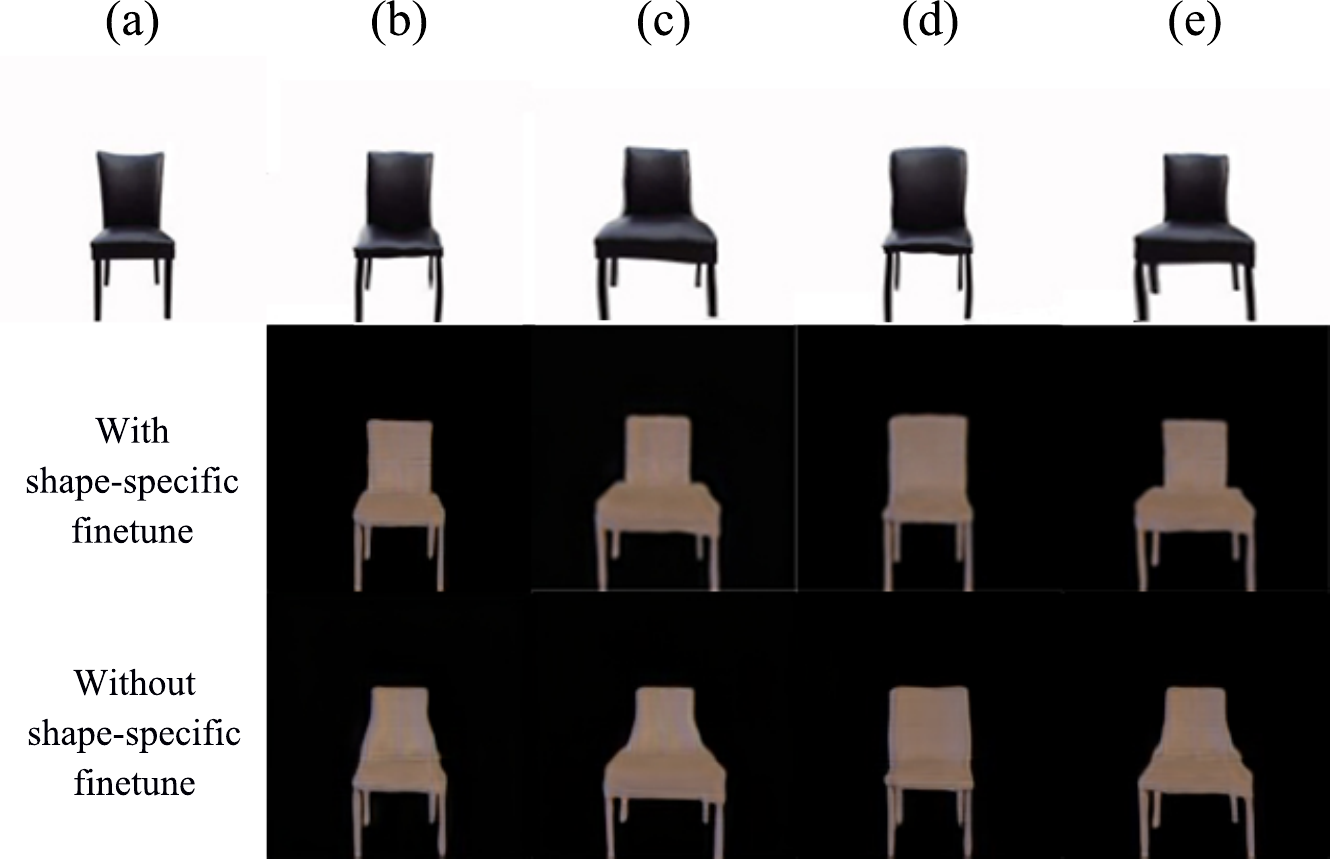}
   \caption{
   \textbf{Editing results of real images.}
   The first row shows the texture mapping results obtained from the shapes inferred through shape-specific finetuning. 
   The second row shows shape images inferred through shape-specific finetune, while the third row shows shape images inferred without shape-specific finetuning.
   (a) Input real image; (b) shape and image reconstructed using our pipeline; (c) seat replacement result; (d) resizing result of the wider back; (e) resizing result of the wider seat.
   }
\label{fig:real_img_fig}
\end{figure}

\subsection{Real image results}
We tested our pipeline for shape part manipulation of real images
To infer the shape attributes of an object in the input real image $I$, we first colorized the shape region in $I$ so that it had the same appearance as our dataset.
We projected the colorized image $\tilde{I}$ into the GAN latent space to obtain its GAN latent code $w_{\tilde{I}}$.
Because the shape in $\tilde{I}$ was often outside the domain covered by our training dataset, our pipeline yielded unsatisfactory results (\textit{without shape-specific finetune} row in \figname~\ref{fig:real_img_fig}).
To address this problem, we designed a shape-specific finetuning process.
We identified the best forward and backward mapping network parameters $\theta_{M_F}'$ and $\theta_{M_B}'$ for $\tilde{I}$ by optimizing the following function:
\begin{align}
\mathcal{L}_{LPIPS}(I', I)+\lambda_3 ||\theta_{M_F}' - \theta_{M_F}||_2 + \lambda_4 ||\theta_{M_B}' - \theta_{M_B}||_2,
\end{align}
where $\theta_{M_F}$ and $\theta_{M_B}$ are the parameters of the pretrained forward and backward mapping functions, and $\lambda_3$ and $\lambda_4$ are the weights of losses representing the distances between the pretrained mapping functions and optimized mapping functions.
\figname~\ref{fig:real_img_fig} shows the manipulated results for a real image obtained through the shape-specific finetuning process.
After obtaining the manipulated image, we warped the input texture in the source real image using thin plate splines (TPS) warping.
We identified the object contours in the input image and the manipulated images using~\cite{suzuki1985topological}, and sampled points on the contours as the control points for TPS warping.
\section{Limitation and future work}
\begin{figure}
\centering
\includegraphics[width=1.0\linewidth]{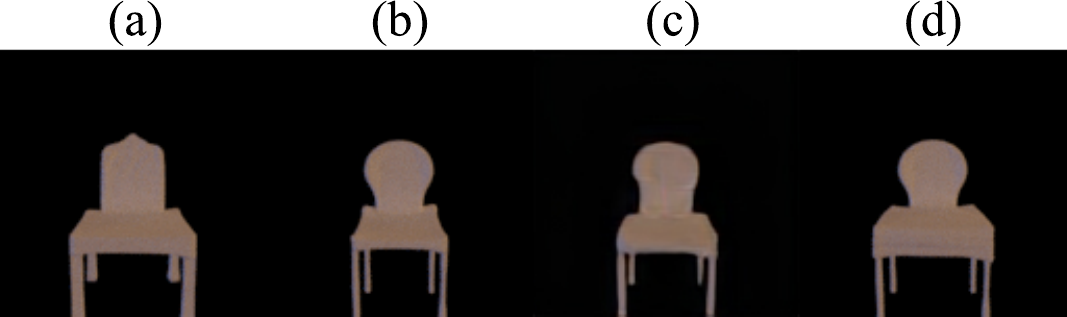}
   \caption{
   (a) The sources of the back and the seat parts\; (b) the source of the leg part; (c) replacement result; (d) ground truth. The replacement result indicates the entanglement between the seat and the leg parts.
   }
\label{fig:entanglement}
\end{figure}
In this paper, we propose a framework for bridging the image latent space and 3D shape attribute space using shape-consistent mapping functions.
Furthermore, we show that the mapping functions enable image shape part manipulation, supporting part replacement, part resizing, and viewpoint manipulation.
However, despite the usefulness of our framework, as demonstrated through manipulations of several man-made object categories, it still has several limitations.
\begin{description}[style=unboxed,leftmargin=0cm]
\item[Entanglement between parts.] 
As shown in \figname~\ref{fig:entanglement}, some non-manipulated parts were altered after part replacement manipulations.
This suggests that some parts were still entangled together in the image latent space.
This could potentially be resolved by using more specific shape part supervision signals, such as part masks.
\item[Fixed shape attribute space.] We trained one end-to-end mapping function for each category and assumed that the number of parts in the input shape was always less than the predefined number.
This limitation is inherited from the adopted shape attribute space we used (\ie~SDMNet).
In the future, we may explore a more flexible shape attribute space.
\item[Gap between synthesized images and real images.] Our framework focuses on the geometry of the input synthesized image.
To manipulate the shape of a real image, we need to infer the appearance of the deformed textures, including from the unseen sides, to achieve realistic manipulation results. 
Moreover, we plan to support the lighting manipulation by incorporating lighting parameters into our framework.
\end{description}
\section{Conclusion}
In this paper, we propose a framework for bridging the image latent space and 3D shape attribute space using shape-consistent mapping functions.
Furthermore, we show that the mapping functions enable image shape part manipulation, supporting part replacement, part resizing, and viewpoint manipulation.

\bibliographystyle{eg-alpha-doi} 
\bibliography{paper}

\newcommand{\etalchar}[1]{$^{#1}$}
\begin{thebibliography}{\uppercase{KAH{\etalchar{*}}20b}}

\bibitem[APCO21]{alaluf2021restyle}
\textsc{Alaluf Y., Patashnik O., Cohen-Or D.}:
\newblock Restyle: A residual-based stylegan encoder via iterative refinement.
\newblock \emph{arXiv preprint arXiv:2104.02699} (2021).

\bibitem[AQW19]{abdal2019image2stylegan}
\textsc{Abdal R., Qin Y., Wonka P.}:
\newblock Image2stylegan: How to embed images into the stylegan latent space?
\newblock In \emph{Proceedings of the IEEE international conference on computer
  vision} (2019), pp.~4432--4441.

\bibitem[AQW20]{abdal2020image2stylegan++}
\textsc{Abdal R., Qin Y., Wonka P.}:
\newblock Image2stylegan++: How to edit the embedded images?
\newblock In \emph{Proceedings of the IEEE/CVF Conference on Computer Vision
  and Pattern Recognition} (2020), pp.~8296--8305.

\bibitem[CB18]{creswell2018inverting}
\textsc{Creswell A., Bharath A.~A.}:
\newblock Inverting the generator of a generative adversarial network.
\newblock \emph{IEEE transactions on neural networks and learning systems 30},
  7 (2018), 1967--1974.

\bibitem[CZS{\etalchar{*}}13]{chen20133}
\textsc{Chen T., Zhu Z., Shamir A., Hu S.-M., Cohen-Or D.}:
\newblock 3-sweep: Extracting editable objects from a single photo.
\newblock \emph{ACM Transactions on Graphics (TOG) 32}, 6 (2013), 1--10.

\bibitem[DTM96]{Paul:1996}
\textsc{Debevec P.~E., Taylor C.~J., Malik J.}:
\newblock Modeling and rendering architecture from photographs: A hybrid
  geometry- and image-based approach.
\newblock In \emph{Proceedings of the 23rd Annual Conference on Computer
  Graphics and Interactive Techniques} (New York, NY, USA, 1996), SIGGRAPH '96,
  Association for Computing Machinery, p.~11–20.
\newblock URL: \url{https://doi.org/10.1145/237170.237191}, \href
  {https://doi.org/10.1145/237170.237191} {\path{doi:10.1145/237170.237191}}.

\bibitem[FFBB21]{Feng:SIGGRAPH:2021}
\textsc{Feng Y., Feng H., Black M.~J., Bolkart T.}:
\newblock Learning an animatable detailed {3D} face model from in-the-wild
  images.
\newblock \emph{ACM Transactions on Graphics (ToG), Proc. SIGGRAPH 40}, 4 (Aug.
  2021), 88:1--88:13.

\bibitem[GAOI19]{Goetschalckx2019GANalyze}
\textsc{Goetschalckx L., Andonian A., Oliva A., Isola P.}:
\newblock Ganalyze: Toward visual definitions of cognitive image properties.
\newblock \emph{arXiv preprint arXiv:1906.10112} (2019).

\bibitem[GG19]{meshrcnn}
\textsc{Georgia~Gkioxari Jitendra~Malik J.~J.}:
\newblock Mesh r-cnn.
\newblock \emph{ICCV 2019} (2019).

\bibitem[GLY{\etalchar{*}}19]{gao2019sparse}
\textsc{Gao L., Lai Y.-K., Yang J., Ling-Xiao Z., Xia S., Kobbelt L.}:
\newblock Sparse data driven mesh deformation.
\newblock \emph{IEEE transactions on visualization and computer graphics}
  (2019).

\bibitem[GSH{\etalchar{*}}19]{Garon_2019_CVPR}
\textsc{Garon M., Sunkavalli K., Hadap S., Carr N., Lalonde J.-F.}:
\newblock Fast spatially-varying indoor lighting estimation.
\newblock In \emph{The IEEE Conference on Computer Vision and Pattern
  Recognition (CVPR)} (June 2019).

\bibitem[GSZ20]{gu2020image}
\textsc{Gu J., Shen Y., Zhou B.}:
\newblock Image processing using multi-code gan prior.
\newblock In \emph{CVPR} (2020).

\bibitem[GTN{\etalchar{*}}20]{guan2020collaborative}
\textsc{Guan S., Tai Y., Ni B., Zhu F., Huang F., Yang X.}:
\newblock Collaborative learning for faster stylegan embedding.
\newblock \emph{arXiv preprint arXiv:2007.01758} (2020).

\bibitem[GYW{\etalchar{*}}19]{gao2019sdm}
\textsc{Gao L., Yang J., Wu T., Yuan Y.-J., Fu H., Lai Y.-K., Zhang H.}:
\newblock Sdm-net: Deep generative network for structured deformable mesh.
\newblock \emph{ACM Transactions on Graphics (TOG) 38}, 6 (2019), 1--15.

\bibitem[HGDG17]{He_2017_ICCV}
\textsc{He K., Gkioxari G., Dollar P., Girshick R.}:
\newblock Mask r-cnn.
\newblock In \emph{Proceedings of the IEEE International Conference on Computer
  Vision (ICCV)} (Oct 2017).

\bibitem[HHLP20]{harkonen2020ganspace}
\textsc{H{\"a}rk{\"o}nen E., Hertzmann A., Lehtinen J., Paris S.}:
\newblock Ganspace: Discovering interpretable gan controls.
\newblock \emph{arXiv preprint arXiv:2004.02546} (2020).

\bibitem[HSSI21]{chong2020ganui}
\textsc{Hin T. C.~L., Shen I.-C., Sato I., Igarashi T.}:
\newblock Interactive optimization of generative image modeling using
  sequential subspace search and content-based guidance.
\newblock \emph{Computer Graphics Forum} (2021).
\newblock \href {https://doi.org/10.1111/cgf.14188}
  {\path{doi:10.1111/cgf.14188}}.

\bibitem[HZZ{\etalchar{*}}20]{huh2020transforming}
\textsc{Huh M., Zhang R., Zhu J.-Y., Paris S., Hertzmann A.}:
\newblock Transforming and projecting images into class-conditional generative
  networks.
\newblock In \emph{European Conference on Computer Vision} (2020), Springer,
  pp.~17--34.

\bibitem[JCI19]{jahanian2019steerability}
\textsc{Jahanian A., Chai L., Isola P.}:
\newblock On the" steerability" of generative adversarial networks.
\newblock \emph{arXiv preprint arXiv:1907.07171} (2019).

\bibitem[JCI20]{gansteerability}
\textsc{Jahanian A., Chai L., Isola P.}:
\newblock On the "steerability" of generative adversarial networks.
\newblock In \emph{International Conference on Learning Representations}
  (2020).

\bibitem[KAH{\etalchar{*}}20a]{karras2020training}
\textsc{Karras T., Aittala M., Hellsten J., Laine S., Lehtinen J., Aila T.}:
\newblock Training generative adversarial networks with limited data.
\newblock \emph{arXiv preprint arXiv:2006.06676} (2020).

\bibitem[KAH{\etalchar{*}}20b]{Karras2020ada}
\textsc{Karras T., Aittala M., Hellsten J., Laine S., Lehtinen J., Aila T.}:
\newblock Training generative adversarial networks with limited data.
\newblock In \emph{Proc. NeurIPS} (2020).

\bibitem[KB14]{kingma2014adam}
\textsc{Kingma D.~P., Ba J.}:
\newblock Adam: A method for stochastic optimization.
\newblock \emph{arXiv preprint arXiv:1412.6980} (2014).

\bibitem[KLA19]{karras2019stylegan}
\textsc{Karras T., Laine S., Aila T.}:
\newblock A style-based generator architecture for generative adversarial
  networks.
\newblock In \emph{Proceedings of the IEEE/CVF Conference on Computer Vision
  and Pattern Recognition} (2019), pp.~4401--4410.

\bibitem[KLA{\etalchar{*}}20]{karras2020analyzing}
\textsc{Karras T., Laine S., Aittala M., Hellsten J., Lehtinen J., Aila T.}:
\newblock Analyzing and improving the image quality of stylegan.
\newblock In \emph{Proceedings of the IEEE/CVF Conference on Computer Vision
  and Pattern Recognition} (2020), pp.~8110--8119.

\bibitem[KSES14]{OM3D2014}
\textsc{Kholgade N., Simon T., Efros A., Sheikh Y.}:
\newblock 3d object manipulation in a single photograph using stock 3d models.
\newblock \emph{ACM Transactions on Computer Graphics 33}, 4 (2014).

\bibitem[LCY{\etalchar{*}}17]{material2017}
\textsc{Liu G., Ceylan D., Yumer E., Yang J., Lien J.-M.}:
\newblock Material editing using a physically based rendering network.
\newblock \emph{International Conference on Computer Vision (ICCV) (spotlight)}
  (2017).

\bibitem[LSG{\etalchar{*}}20]{pix2surf_2020}
\textsc{Lei J., Sridhar S., Guerrero P., Sung M., Mitra N., Guibas L.~J.}:
\newblock Pix2surf: Learning parametric 3d surface models of objects from
  images.
\newblock In \emph{Proceedings of European Conference on Computer Vision
  ({ECCV})} (2020).
\newblock URL: \url{https://geometry.stanford.edu/projects/pix2surf}.

\bibitem[MDG{\etalchar{*}}20]{mao2020std}
\textsc{Mao A., Dai C., Gao L., He Y., Liu Y.-j.}:
\newblock Std-net: Structure-preserving and topology-adaptive deformation
  network for 3d reconstruction from a single image.
\newblock \emph{arXiv preprint arXiv:2003.03551} (2020).

\bibitem[MGY{\etalchar{*}}19]{mo2019structurenet}
\textsc{Mo K., Guerrero P., Yi L., Su H., Wonka P., Mitra N., Guibas L.}:
\newblock Structurenet: Hierarchical graph networks for 3d shape generation.
\newblock \emph{ACM Transactions on Graphics (TOG), Siggraph Asia 2019 38}, 6
  (2019), Article 242.

\bibitem[PBH20]{plumerault2020controlling}
\textsc{Plumerault A., Borgne H.~L., Hudelot C.}:
\newblock Controlling generative models with continuous factors of variations.
\newblock \emph{arXiv preprint arXiv:2001.10238} (2020).

\bibitem[PGM{\etalchar{*}}19]{NEURIPS2019_9015}
\textsc{Paszke A., Gross S., Massa F., Lerer A., Bradbury J., Chanan G.,
  Killeen T., Lin Z., Gimelshein N., Antiga L., Desmaison A., Kopf A., Yang E.,
  DeVito Z., Raison M., Tejani A., Chilamkurthy S., Steiner B., Fang L., Bai
  J., Chintala S.}:
\newblock Pytorch: An imperative style, high-performance deep learning library.
\newblock In \emph{Advances in Neural Information Processing Systems 32},
  Wallach H., Larochelle H., Beygelzimer A., d\textquotesingle Alch\'{e}-Buc
  F., Fox E., Garnett R., (Eds.). Curran Associates, Inc., 2019,
  pp.~8024--8035.
\newblock URL:
  \url{http://papers.neurips.cc/paper/9015-pytorch-an-imperative-style-high-performance-deep-learning-library.pdf}.

\bibitem[S{\etalchar{*}}85]{suzuki1985topological}
\textsc{Suzuki S., et~al.}:
\newblock Topological structural analysis of digitized binary images by border
  following.
\newblock \emph{Computer vision, graphics, and image processing 30}, 1 (1985),
  32--46.

\bibitem[SEBM20]{spingarn2020gan}
\textsc{Spingarn-Eliezer N., Banner R., Michaeli T.}:
\newblock Gan" steerability" without optimization.
\newblock \emph{arXiv preprint arXiv:2012.05328} (2020).

\bibitem[SFG17]{su2017a}
\textsc{Su H., Fan H., Guibas L.}:
\newblock {A} {P}oint {S}et {G}eneration {N}etwork for 3{D} {O}bject
  {R}econstruction from a {S}ingle {I}mage.
\newblock In \emph{The IEEE Conference on Computer Vision and Pattern
  Recognition (CVPR)} (2017).

\bibitem[SGTZ20]{shen2020interpreting}
\textsc{Shen Y., Gu J., Tang X., Zhou B.}:
\newblock Interpreting the latent space of gans for semantic face editing.
\newblock In \emph{Proceedings of the IEEE/CVF Conference on Computer Vision
  and Pattern Recognition} (2020), pp.~9243--9252.

\bibitem[SSH20]{Song_2020_CVPR}
\textsc{Song C., Song J., Huang Q.}:
\newblock Hybridpose: 6d object pose estimation under hybrid representations.
\newblock In \emph{IEEE/CVF Conference on Computer Vision and Pattern
  Recognition (CVPR)} (June 2020).

\bibitem[SSS06]{SSS:2006}
\textsc{Snavely N., Seitz S.~M., Szeliski R.}:
\newblock Photo tourism: Exploring photo collections in 3d.
\newblock In \emph{SIGGRAPH Conference Proceedings} (New York, NY, USA, 2006),
  ACM Press, pp.~835--846.

\bibitem[SZ21]{shen2021closedform}
\textsc{Shen Y., Zhou B.}:
\newblock Closed-form factorization of latent semantics in gans.
\newblock In \emph{CVPR} (2021).

\bibitem[TAN{\etalchar{*}}21]{tov2021designing}
\textsc{Tov O., Alaluf Y., Nitzan Y., Patashnik O., Cohen-Or D.}:
\newblock Designing an encoder for stylegan image manipulation.
\newblock \emph{ACM Transactions on Graphics (TOG) 40}, 4 (2021), 1--14.

\bibitem[TEB{\etalchar{*}}20a]{tewari2020pie}
\textsc{Tewari A., Elgharib M., Bernard F., Seidel H.-P., P{\'e}rez P.,
  Zollh{\"o}fer M., Theobalt C.}:
\newblock Pie: Portrait image embedding for semantic control.
\newblock \emph{ACM Transactions on Graphics (TOG) 39}, 6 (2020), 1--14.

\bibitem[TEB{\etalchar{*}}20b]{tewari2020stylerig}
\textsc{Tewari A., Elgharib M., Bharaj G., Bernard F., Seidel H.-P., P{\'e}rez
  P., Zollhofer M., Theobalt C.}:
\newblock Stylerig: Rigging stylegan for 3d control over portrait images.
\newblock In \emph{Proceedings of the IEEE/CVF Conference on Computer Vision
  and Pattern Recognition} (2020), pp.~6142--6151.

\bibitem[TZK{\etalchar{*}}17]{tewari2017mofa}
\textsc{Tewari A., Zollhofer M., Kim H., Garrido P., Bernard F., Perez P.,
  Theobalt C.}:
\newblock Mofa: Model-based deep convolutional face autoencoder for
  unsupervised monocular reconstruction.
\newblock In \emph{Proceedings of the IEEE International Conference on Computer
  Vision Workshops} (2017), pp.~1274--1283.

\bibitem[WCH{\etalchar{*}}21]{wang2021cross}
\textsc{Wang C., Chai M., He M., Chen D., Liao J.}:
\newblock Cross-domain and disentangled face manipulation with 3d guidance.
\newblock \emph{arXiv preprint arXiv:2104.11228} (2021).

\bibitem[WLS21]{wu2021stylespace}
\textsc{Wu Z., Lischinski D., Shechtman E.}:
\newblock Stylespace analysis: Disentangled controls for stylegan image
  generation.
\newblock In \emph{Proceedings of the IEEE/CVF Conference on Computer Vision
  and Pattern Recognition} (2021), pp.~12863--12872.

\bibitem[WP21]{wang2021geometry}
\textsc{Wang B., Ponce C.~R.}:
\newblock The geometry of deep generative image models and its applications.
\newblock \emph{arXiv preprint arXiv:2101.06006} (2021).

\bibitem[WWX{\etalchar{*}}17]{marrnet}
\textsc{Wu J., Wang Y., Xue T., Sun X., Freeman W.~T., Tenenbaum J.~B.}:
\newblock {MarrNet: 3D Shape Reconstruction via 2.5D Sketches}.
\newblock In \emph{Advances In Neural Information Processing Systems} (2017).

\bibitem[YYTK21]{Yu_2021_CVPR}
\textsc{Yu A., Ye V., Tancik M., Kanazawa A.}:
\newblock pixelnerf: Neural radiance fields from one or few images.
\newblock In \emph{Proceedings of the IEEE/CVF Conference on Computer Vision
  and Pattern Recognition (CVPR)} (June 2021), pp.~4578--4587.

\bibitem[YYY{\etalchar{*}}16]{NIPS2016_6206}
\textsc{Yan X., Yang J., Yumer E., Guo Y., Lee H.}:
\newblock Perspective transformer nets: Learning single-view 3d object
  reconstruction without 3d supervision.
\newblock In \emph{Advances in Neural Information Processing Systems 29}, Lee
  D.~D., Sugiyama M., Luxburg U.~V., Guyon I., Garnett R., (Eds.). Curran
  Associates, Inc., 2016, pp.~1696--1704.
\newblock URL:
  \url{http://papers.nips.cc/paper/6206-perspective-transformer-nets-learning-single-view-3d-object-reconstruction-without-3d-supervision.pdf}.

\bibitem[ZKSE16]{zhu2016generative}
\textsc{Zhu J.-Y., Kr{\"a}henb{\"u}hl P., Shechtman E., Efros A.~A.}:
\newblock Generative visual manipulation on the natural image manifold.
\newblock In \emph{Proceedings of European Conference on Computer Vision
  (ECCV)} (2016).

\bibitem[ZSZZ20]{zhu2020indomain}
\textsc{Zhu J., Shen Y., Zhao D., Zhou B.}:
\newblock In-domain gan inversion for real image editing.
\newblock In \emph{Proceedings of European Conference on Computer Vision
  (ECCV)} (2020).

\bibitem[ZZLS21]{Zhu_2021_CVPR}
\textsc{Zhu Y., Zhang Y., Li S., Shi B.}:
\newblock Spatially-varying outdoor lighting estimation from intrinsics.
\newblock In \emph{Proceedings of the IEEE/CVF Conference on Computer Vision
  and Pattern Recognition (CVPR)} (June 2021), pp.~12834--12842.

\end{thebibliography}

\end{document}